\documentclass[conference]{IEEEtran}
\IEEEoverridecommandlockouts
\usepackage{cite}
\usepackage{amsmath,amssymb,amsfonts}
\usepackage{algorithmic}
\usepackage{graphicx}
\usepackage{textcomp}
\usepackage{xcolor}
\def\BibTeX{{\rm B\kern-.05em{\sc i\kern-.025em b}\kern-.08em
    T\kern-.1667em\lower.7ex\hbox{E}\kern-.125emX}}
\begin{document}

\title{Generative-Adversarial Networks for Low-Resource Language Data
Augmentation in Machine Translation\\}

\author{\IEEEauthorblockN{Linda Zeng}
\IEEEauthorblockA{
\textit{The Harker School}\\
San Jose, United States of America \\
26lindaz@students.harker.org}
~\\
}

\maketitle

\begin{abstract}
Neural Machine Translation (NMT) systems struggle when translating to and from low-resource languages, which lack large-scale data corpora for models to use for training. As manual data curation is expensive and time-consuming, we propose utilizing a generative-adversarial network (GAN) to augment low-resource language data. When training on a very small amount of language data (under 20,000 sentences) in a simulated low-resource setting, our model shows potential at data augmentation, generating monolingual language data with sentences such as “ask me that healthy lunch im cooking up,” and “my grandfather work harder than your grandfather before.” Our novel data augmentation approach takes the first step in investigating the capability of GANs in low-resource NMT, and our results suggest that there is promise for future extension of GANs to low-resource NMT.
\end{abstract}


\section{Introduction}
Although technology has become a staple of daily life, society’s best-performing computing systems still fail to reflect the world’s diversity in languages. Current translation models frequently err when translating to and from “low-resource languages” \cite{gu-etal-2018-universal}, which are languages that do not have much digital data that the machine learning algorithms underlying the software can use as reference.

While vast and detailed language data exist for high-resource languages such as English and Spanish, low-resource languages, including many American indigenous languages like Aymara and Quechua \cite{zheng-etal-2021-low}, lack large-scale corpora to be used for training. Because models learn the syntactic and lexical patterns underlying translations through processing training data, an insufficient amount of data hinders them from producing accurate translations, and consequently, models often generate incorrect translations for low-resource languages \cite{zoph-etal-2016-transfer}. 

Prior research has addressed this problem, but few truly solve the issue. Previous approaches focus on transferring learning between high-resource and low-resource languages \cite{gu-etal-2018-universal, zoph-etal-2016-transfer, zheng-etal-2021-low}, which have limited efficacy depending on the similarity between the high-resource and low-resource languages being used. The direction of data augmentation with completely original sentences has not been fully explored \cite{fadaee-etal-2017-data} and holds promise for breakthrough, as data augmentation directly addresses the challenge of lacking training data. Both monolingual and parallel data augmentation is important for Neural Machine Translation (NMT), as training on monolingual corpora in addition to parallel data has been used to improve NMT models \cite{zhang-zong-2016-exploiting, sennrich-etal-2016-improving, cai-etal-2021-neural}, especially in low-resource NMT \cite{currey-etal-2017-copied}, and our system performs monolingual data augmentation.

In contrast to the human labor of creating new sentences in low-resource languages by hand, a generative-adversarial network (GAN) is capable of autonomously generating unlimited amounts of new data. \cite{yang-etal-2018-improving, zhang-etal-2018-bidirectional, yang-etal-2018-unsupervised, rashid-etal-2019-bilingual} have implemented GANs for general NMT, but at the time of our research, no previous models have used them for text augmentation of low-resource languages.

Using a simulated low-resource setting of only 20,000 training data points, we explore the capability of a GAN for monolingual low-resource language data augmentation to improve machine translation quality. We build on the structure of the GAN from \cite{yang-etal-2018-improving}, which directly translates between high-resource languages, to generate a completely new low-resource language corpus from noise. Overall, our research is the first to combine GANs, data augmentation, and low-resource NMT. 
\section{Related Work}

\subsection{Preliminaries on NMT}

NMT uses neural networks to translate between two different languages and commonly uses an encoder-decoder architecture \cite{kalchbrenner-blunsom-2013-recurrent, cho-etal-2014-learning, sutskever2014sequence}. A sequence-to-sequence encoder-decoder \cite{cho-etal-2014-learning} traditionally uses two recurrent neural networks (RNNs), called the encoder and the decoder. The encoder converts a sentence in a given language into latent space (encoding) while the decoder takes the latent space and converts it back into a sentence in the other language (decoding). As the latent space representations, also known as embeddings, represent the core meaning of the sentence, the output of the decoder is a direct translation of the sentence that was input to the encoder. Long-short term memory (LSTM) networks \cite{10.1162/neco.1997.9.8.1735} are a type of RNN that can operate on a sequence of words and are commonly used in NMT because they can capture long-term dependencies between sequential data points. However, because they evaluate from left to right, the encoder-decoder does not examine the context that appears after a word. As a result, these networks require multiple instances of words appearing in diverse contexts in order to create vectors that accurately represent the context needed for these words in the latent space \cite{fadaee-etal-2017-data}, causing NMT to frequently err with low-resource pairs \cite{zoph-etal-2016-transfer}. NMT models based on Transformers \cite{vaswani2017attention} have also risen in popularity, as the Transformer framework uses attention to improve parallelizability of training. After implementing both the RNNSearch and the Transformer on their GAN model, \cite{yang-etal-2018-improving} found that both architectures could be applied with similar performances. 

\subsection{Data Augmentation for Low-Resource NMT}
Other models and software have tried to tackle this issue by using multilingual transfer-learning approaches \cite{zheng-etal-2021-low, gu-etal-2018-universal, zoph-etal-2016-transfer} and a word substitution approach \cite{fadaee-etal-2017-data}. The transfer-learning approach teaches a model to use its knowledge from high-resource language pairs and apply it to low-resource languages, but it involves finding high resource languages that are very close to the language at hand, which is difficult depending on the language family. While the Germanic language family has more related high-resource languages such as German and Danish \cite{tchistiakova2021edinsaarwmt21, chen2021machine}, African languages' closest high-resource languages are English and French \cite{dione-2021-multilingual}. Conversely, the data augmentation method in \cite{fadaee-etal-2017-data} involves altering translation data by replacing specific words in given sentences, thereby diversifying the context in which these words show up. However, they do not generate completely new sentences to be used for data augmentation. As a result, while the model learns contexts for individual words, the model still suffers from lack of diverse grammatical structures and sentence topics from which it can learn how to structure full sentences.

\subsection{Preliminaries on GANs}
Commonly used in image generation and computer vision \cite{yang-etal-2018-improving}, GANs combine two machine learning models \cite{goodfellow2014generative}. The first one is called the generator, a model that takes in input and generates samples, which are then fed into the discriminator, the second machine learning model. The discriminator is given samples either from the real data or from the generator, and it must determine if this sample is real or generated. If it correctly predicts, this indicates that it is improving. If it incorrectly predicts, this indicates that the generator is improving. Then, depending on how well the discriminator predicted, both models tweak their weights to continue this cycle. The generator “wins” when the discriminator cannot tell the two samples apart, and the discriminator “wins” if it can tell them apart. Both models want to win against the other, so each one keeps improving by learning from the other until they finally reach equilibrium, where both models are performing optimally.

\subsection{GANs in NLP}
In the past, \cite{betti-etal-2020-controlled} and \cite{ahamad-2019-generating} were successful in generating synthetic text through adversarial training, and \cite{elneima-binkowski-2022-adversarial} implemented a GAN for low-resource speech augmentation for text-to-speech. To our knowledge, \cite{yang-etal-2018-improving} was the first to use GANs for machine translation, and we base our model on theirs. Their generator learned to translate a source-language sentence into its target-language translation while the discriminator tried to distinguish between real and generated translations. Continuing with this research, they implemented two GANs to use as tools to ensure the efficacy of their encoder models in weight sharing for latent space embeddings \cite{yang-etal-2018-unsupervised}. \cite{zhang-etal-2018-bidirectional} introduced bidirectional GANs to improve translation quality by creating another generator model to act as the discriminator, and the \cite{rashid-etal-2019-bilingual} study used a latent space based GAN to translate bidirectionally in both a supervised and unsupervised setting. To our knowledge, \cite{kumar2023exploiting} is the only other work that has applied GANs to low-resource language NMT, and their work was conducted in parallel to ours. They use GANs for direct translation rather than data augmentation.

\section{Model Architecture}

\subsection{Overall Workflow}
\begin{figure}[h]
	\includegraphics[width=\linewidth]{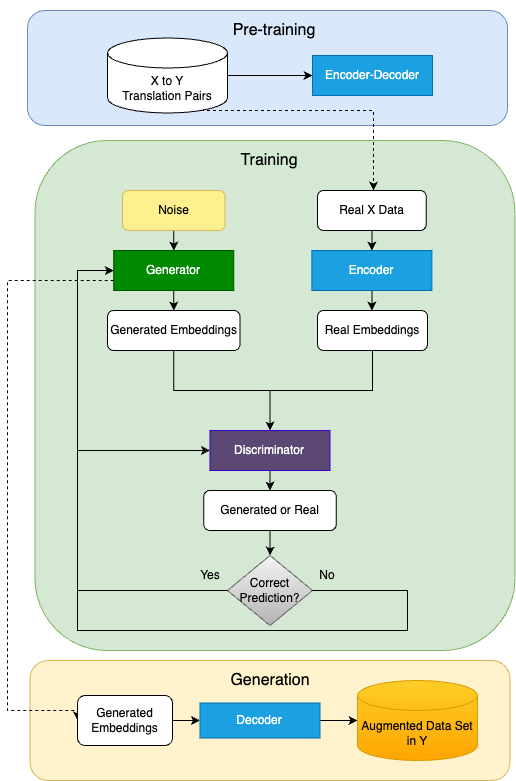}
	\caption{Overall Workflow}
	\label{fig:workflow}
\end{figure}

Our model workflow (shown in Fig.~\ref{fig:workflow}) includes three stages: 1) pre-training of the encoder-decoder, 2) training of the GAN, and 3) generation of the augmented data set. 

In the first stage, the encoder-decoder is trained on the human-created parallel corpus and learns to translate from language X to language Y. The encoder encodes X data into latent space embeddings while the decodes the embeddings into the corresponding translations in Y. The Y data are compared against the reference translations in the corpus, and the errors are backward propagated. The model repeatedly adjusts its weights until it is performing optimally.

In the second stage, the encoder-decoder's weights are frozen to train the GAN. The generator takes in a batch of random noise (randomly generated numbers between -1 to 1 which by themselves had no meaning), and attempts to assign meaning to them by rearranging them into latent space embeddings. A batch of sentences in language X are fed into the encoder, and it encodes them into “real” latent space embeddings. Both the generated and real embeddings are given as input into the discriminator, which classifies a given embedding as generated or real.

The discriminator's prediction is compared with the actual label of the embedding, and its errors are backpropagated up to the generator. 
When the discriminator classifies correctly, it is rewarded, and when it classifies incorrectly, meaning it cannot differentiate between the two embeddings, the generator is rewarded. Consequently, despite not having direct access to the encoder's embeddings, the generator learns to generate embeddings closer to the encoder's embeddings because it is randomly guessing until it  awarded each time it does so. Through trial and error, the generator learns to create encodings more similar to the encoder’s in order to fool the discriminator. 

In the third stage, no training is involved. Once the GAN is performing optimally, the generator is given noise and generates a large corpus of embeddings. The decoder decodes these embeddings into sentences in language Y that correspond to the embeddings' meanings in latent space. These sentences form the newly-augmented monolingual data corpus in language Y. The generator can be run an infinite amount of times to generate as much data as necessary.

It is important to note that while most GAN architectures use sentences straight from the reference corpus as reference or "real" data, we use the encoder's embedding outputs as the reference data. As a result, our generator learns to generate latent space embeddings rather than immediately intelligible sentences in specific languages. To generalize this data augmentation method to other languages, only the encoder-decoder needs to be retrained, so that it learns to embed different languages.

\subsection{Underlying Architectures}

\begin{figure*}
    \centering
     \includegraphics[width=1.0\textwidth]{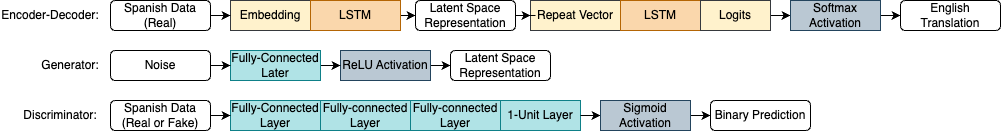}
	\caption{Model Architectures}
	\label{fig:architecture}
\end{figure*}

Shown in Fig.~\ref{fig:architecture}, the encoder-decoder uses bidirectional LSTMs. First, an embedding layer learns to map words with analogous meanings to similar numerical vectors and inputs them into the encoder LSTM, which includes weight decay and dropout to minimize overfitting. The latent-space output of the encoder LSTM is copied as input for the decoder LSTM, using a repeat vector. The decoder LSTM includes weight decay and dropout and decodes the latent space representations into numerical outputs. The logits layer maps the numerical outputs of the decoder LSTM into vectors that correspond with each word. The model uses a softmax activation function and is optimized with a categorical cross entropy loss function and an Adam optimizer \cite{kingma2017adam}.

Shown in Fig.~\ref{fig:architecture}, the generator consists of a single fully-connected layer and a reLu activation function. It is optimized with a categorical cross entropy loss function and an Adam optimizer. The loss is calculated based on the final loss value of the GAN and is inversely proportional to the loss of the discriminator.

Shown in Fig.~\ref{fig:architecture}, the discriminator consists of three fully-connected layers, followed by a one-unit dense layer and a sigmoid activation function, which produces a prediction of whether the input is from the encoder or the generator. It is optimized with a binary cross entropy loss function and an Adam optimizer. The loss is calculated based on the final loss value of the GAN.

The GAN consists of the initialized generator, followed by the discriminator. The input of the GAN is noise, which is fed into the generator. The output of the generator is fed into the discriminator, and the output of the discriminator is the prediction, which is used for backpropagation through the entire system to calculate loss values for the generator and discriminator. It is trained using a binary cross entropy loss function and an Adam optimizer.

Hyperparameter values and loss values during training are included in Appendix \ref{section:implement} and Appendix \ref{section:training}, respectively.
\section{Data}

\subsection{Simulated Low-Resource Setting}
In order to investigate the GAN model’s ability to augment low-resource language data, we mimic the characteristics of low-resource languages in English and Spanish. In the English to Spanish data set, we reduced the amount of data and used only 20,000 of the data set’s 253,726 sentence pairs to train and test the model. As a result, we were able to replicate the effect on low-resource languages and evaluate the model’s effectiveness after it was trained on 20,000 sentences.

\subsection{Training Data}

The English to Spanish data set we used to train the model is from Tatoeba\footnote{https://tatoeba.org/en/} and has been pre-processed and cleaned by a third party\footnote{http://www.manythings.org/} who downloaded the original file from Tatoeba. Founded by Trang Ho in 2006, Tatoeba is a database containing collections of parallel translations contributed by thousands of members. It offers data for 419 languages that are easy to download. 19,000 sentence pairs of the 20,000 in the dataset were used to train the encoder-decoder.

Within the 19,000 sentences of training data, 1,000 sentence pairs were separated to be used as a test during each epoch of training to display validation accuracy and indicate overfitting of the encoder-decoder.

\subsection{Test Data}

The last 1,000 sentence pairs in the 20,000 sentence English to Spanish data set from Tatoeba were not seen by the encoder-decoder and were used as test data, given to the encoder-decoder after it finished training to evaluate the model's efficacy.

Numerical characteristics of both the training and test data are included in Appendix \ref{section:characteristics}.

\subsection {Preprocessing}
We began by removing punctuation from the language data and converting it to lowercase in order to standardize it and retain only the words themselves. We then tokenized the sentence data using the Keras Tokenizer, which split each sentence into a list of probabilities representing its constituent words. Because not all of the sentences had the same number of words, we found the longest sentence and padded the rest with zeros so that every sentence would be the same length. 

\section{Results}

\subsection{Encoder-Decoder Performance} \label{section:Encoder-DecoderPerformance}
On the test data, the encoder-decoder had a final accuracy of 69.3\%. We accept this accuracy under two considerations. First, the model was trained on less than 20,000 sentences, less than one-tenth of the size of other low-resource language data sets, whose sizes are between 0.2 to 1 million \cite{ranathunga2021neural, zoph-etal-2016-transfer}. Second, the encoder-decoder is not the final model we propose, as it serves only as a guide for our GAN to learn how to generate latent space representations. As this paper focuses on the data augmentation aspect of our model, accuracy is used to give a high-level indication of the encoder-decoder's performance so that we may continue on to training the GAN.

In the future, a pre-trained encoder-decoder may be used to stream-line the process, and as the research shifts from the data generation task to the end-to-end machine translation task, further evaluation, including BLEU scores \cite{papineni-etal-2002-bleu}, will be performed.

\subsection{GAN Performance} \label{section:GANPerformance}

The generator was able to generate successful, coherent sentences, shown in Table \ref{table:results}. Upon manual inspection, we found that the model was able to reproduce both syntactic and semantic meaning in some cases, although the model displayed significant errors in many other cases. Analysis of the errors made and limitations is discussed further below. Syntactically, the model placed words in the correct places based on their part of speech. A majority of the sentences also centered around a cohesive theme, indicating the model’s successful understanding of word meanings, and communicated a specific message. From random noise, the generator was able to create its own completely new and logical sentences, a significant feat considering the lack of training data.

\begin{table*}
\centering
\caption{\label{table:results}
Sample Sentences Generated by the GAN
}
\begin{tabular}{lll}
\hline
\textbf{Sample Generated Sentences} & \textbf{Qualitative Evaluation} \\
\hline
\ my grandfather work harder than your grandfather before & good\\
\ to consider quit job is this dream man & good\\
\ ask me that healthy lunch im cooking up & good\\
\ maryam discovered hes hes am am are are & repetition\\
\ home actually was everything everything listen actually everything & repetition\\
\ cheerful weird yourself punished music alone everybody everybody & nonsensical\\
\ those in so friends so complicated english comes & nonsensical\\
\ stressed gloves eating eating worried online online online & unrelated\\
\hline
\end{tabular}
\end{table*}

It remains a challenge to quantitatively evaluate synthetically-generated data due to the lack of comparable reference sentences, and due to limited resources, we did not qualitatively evaluate every generated sentence.

\subsection{Error Analysis} \label{section:ErrorAnalysis}

While the GAN was successfully able to generate some coherent sentences from scratch, the GAN made a significant number of errors. We qualitatively observed that the severity of the errors decreased as the GAN trained for more epochs. Thus, future models may train for a larger number of epochs to examine whether the frequency and severity of errors are able to reduce significantly further. 

\subsubsection{Repeated Words}

Frequently, the GAN generated sentences with repeating words, shown in Table \ref{table:results}. We hypothesize that this issue, which occurs in most NMT models \cite{fu2021theoretical}, is due to the fact that the model is trying to generate words that are close in context to each other, which is necessary in order for a sentence to make sense. However, the model may not know or weigh in if it already used a word or not and ends up repeating that same word. It likely tries to find a word that is close in context to the prior word, and because related words have closer probabilities, it generates a probability very close to the previous word’s probability. Then, after the decoder translates the latent space representations into probabilities, the close probabilities may be reduced to the same word because the words with the highest close probabilities are chosen to represent the probabilities the GAN returned. There may not be enough words with intricate probabilities close to a given one, so that one is chosen for all words with probabilities in a certain range around it. \cite{fu2021theoretical} theorize that the issue is in the nature of languages themselves, as some words tend to predict themselves as the next word in context. Some incoherencies in the generated sentences could also be due to the fact the model may not fully understand the grammatical structure of sentences and resorts to repeating words it does know how to represent in order to fill up space. Potential solutions would be to train the model to remember the previous probabilities it generated and to vary its generated probabilities more to ensure that it does not repeat very similar ones.

\subsubsection{Nonsensical Grammar}
Other sentences contained minimal repetition but were still grammatically incorrect or nonsensical, shown in Table \ref{table:results}. We believe that these sentences contain relatively randomly-placed words because the model has not learned about these words with enough context to know where to place them grammatically. Because it has seen these well-known yet complex words, it can generate them but is unsure of how many to generate, where to place them, or which words to surround them with. The model also may have mistakenly generated words with similar probabilities in search of words with close context. For example, “cheerful” and “weird” are both adjectives that describe “yourself,” so their probabilities may be similar enough for the model to generate them together. However, the model does not understand that these words have parallel meanings, and that only one should be used. A possible avenue for future work is to explore training the model on the difference between probabilities of words parallel in meaning and probabilities of related words that are required to be together in order to form a sentence.

\subsubsection{Unrelated Words}
Although the model generally uses related words like “studies” and “novels” together, it occasionally groups unrelated words together, shown in Table \ref{table:results}. We hypothesize that this is because it has not seen a word (like “gloves”) enough to understand its usual context (being put on people’s hands). However, it does understand that gloves is a noun and has previously seen nouns (such as people) being stressed, eating, worrying, and going online. A potential future path to explore would be incorporating a dictionary of words into the model’s training so that it better understands the words’ meanings.

\section{Conclusion}

Because of its ability to generate an unlimited amount of original sentences despite being trained on minimal data, this GAN architecture can be used as a tool to augment low-resource language data, allowing translation models to train on more sentences in order to generate more accurate translations. This research is the first to apply a GAN to data augmentation in the low-resource NMT task, and we find promising results in cohesiveness and coherency of generated sentences. This work serves as a reference to encourage future work combining GANs and low-resource NMT. Improvements can be made on this research to increase the comprehensiveness of model evaluation and to minimize the repetition and incoherence in many of the generated sentences. One promising future direction is to train the model to understand the previous words it has generated and to remember the grammatical relationship between a word and others of similar probabilities. 

\section{Limitations}

As the model was trained in a simulated low-resource setting, its performance on real low-resource languages would depend on these languages' similarities to English and Spanish. Specifically, isolating languages, which have limited morphology, would work better with this model.

For the encoder-decoder, as the small amount of data lend itself to severe overfitting, further research could be done to minimize overfitting through reducing model capacity, implementing L1 regularization in addition to the current L2 regularization being used, experimenting with stronger dropout, and applying cross-validation.

Thoroughly discussed in Section \ref{section:ErrorAnalysis}, various directions also exist to improve the accuracy and reliability of the GAN's generated sentences, from remembering previous probabilities to training the model to distinguish words parallel in meaning.

As mentioned in Sections \ref{section:Encoder-DecoderPerformance} and \ref{section:GANPerformance}, we plan to include further evaluation. As a next step, we will include BLEU scores for the encoder-decoder. For the GAN, we will use statistical analysis and thorough human evaluation, and a future direction is to use the synthetic data to train an NMT model to improve upon current baselines.

While this model contributes to machine translation by augmenting monolingual data and the use of monolingual corpora is becoming increasingly prevalent in NMT models \cite{cai2021neural}, extending this research to generate parallel translations would allow for a larger impact, as NMT models often train on parallel data in addition to monolingual corpora. Also, future software can be developed to clean the generated sentences from the GAN and extract only the coherent ones in order to add them to data sets. Reinserting punctuation and capitalization serves as another future area for exploration.

\appendices

\section{Data Characteristics}\label{section:characteristics}

Table \ref{tab:trainingdata} and Table \ref{tab:testdata} capture key statistical characteristics of the training and test data, respectively.

\begin{table}[h]
\centering
\caption{Characteristics of Training Data}
\begin{tabular}{lll}
\hline
\textbf{Characteristic} & \textbf{English} & \textbf{Spanish} \\
\hline
\ Average Sentence Length & 4.72 & 4.52 \\
\ Max Sentence length & 8 & 11 \\
\ Mean & 204.38 & 300.87 \\
\ Standard Deviation & 595.44 & 1055.53 \\
\hline
\end{tabular}
\label{tab:trainingdata}
\end{table}

\begin{table}[h]
\centering
\caption{Characteristics of Test Data}
\begin{tabular}{lll}
\hline
\textbf{Characteristic} & \textbf{English} & \textbf{Spanish} \\
\hline
\ Average Sentence Length & 4.71 & 4.53 \\
\ Max Sentence length & 7 & 9 \\
\ Mean & 213.62 & 337.18 \\
\ Standard Deviation & 662.48 & 1285.73 \\
\hline
\end{tabular}
\label{tab:testdata}
\end{table}

\section{Implementation Details}\label{section:implement}
We use layers imported from the Keras Python library \cite{chollet2015keras}. The hyperparameters we used are listed in Table \ref{tab:hyperparameters}.

\begin{table}[b]
\centering
\caption{Hyperparameter Values}
\begin{tabular}{lll}
\hline
\textbf{Hyperparameter} & \textbf{Value} \\
\hline
\ Epochs (Encoder-Decoder) & 400 \\
\ Batch Size (Encoder-Decoder) & 30 \\
\ LSTM Units (Encoder-Decoder) & 256 \\
\ LSTM Dropout (Encoder) & 0.5 \\
\ LSTM Dropout (Decoder) & 0.5 \\
\ Logits Dropout (Encoder-Decoder) & 0.5 \\
\ L2 Regularizer (Encoder) & 5e-5 \\
\ L2 Regularizer (Decoder) & 1e-5 \\
\ Learning Rate (Encoder-Decoder) & 2e-3 \\
\ Beta1 Decay (Encoder-Decoder) & 0.7 \\
\ Beta2 Decay (Encoder-Decoder) & 0.97 \\
\ Epochs (GAN) & 8000 \\
\ Batch Size (GAN) & 1900 \\
\ Learning Rate (GAN) & 1e-4 \\
\ Dense Units (Generator) & 256 \\
\ Learning Rate (Generator) & 4e-4 \\
\ Dense Units (Discriminator) & 1024 \\
\ Learning Rate (Discriminator) & 1e-4 \\
\hline
\end{tabular}
\label{tab:hyperparameters}
\end{table}

In order to find the most optimal hyperparameters, we first varied the values by a factor of either 2 or 10 and tested every combination of the values with each other. To optimize time, we used 5,000 sentences and 80 epochs. For the learning rates of the encoder-decoder, generator, discriminator, and GAN as well as the L2 regularizers, we tried a range of values from 1e-1 to 1e-8 decreasing in magnitude by a factor of 10 each time. When varying the number of units and batch sizes for the encoder-decoder’s LSTM layers, the generator’s dense layer, and the discriminator’s dense layers, we chose powers of 2 between 16 and 2048. For the encoder-decoder’s dropouts, we tried a range from 0.5 to 0.8.

After finding the approximate values to optimize performance, we tested more specific values within the ideal range we found, isolating each of the models and incrementing values by around 1 to 10. We continued this process until we found the most optimal parameters and then increased the amount of training data and epochs.

\section{Training} \label{section:training}
\begin{figure}[h]
\centering
\includegraphics[width=\linewidth]{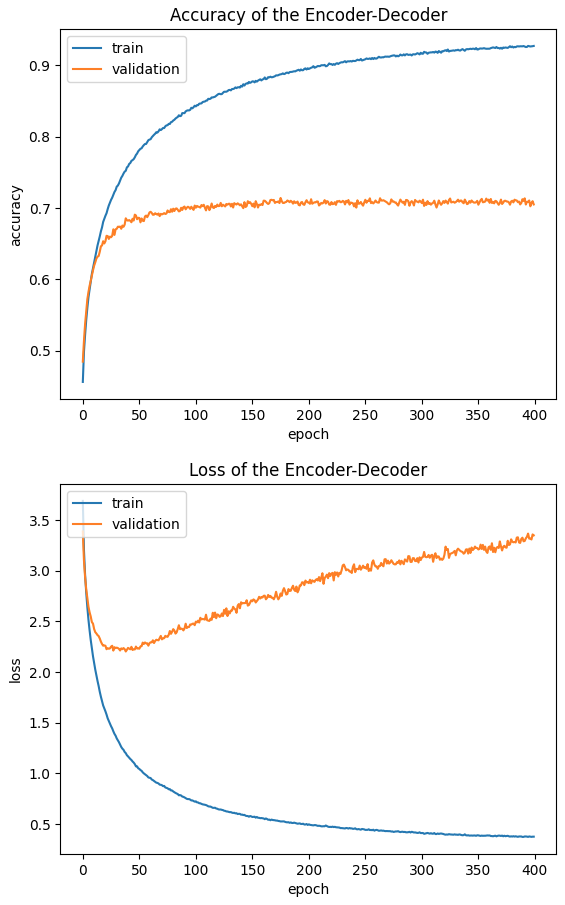}
\caption{Accuracy and Loss of the Encoder-Decoder during Training}
\label{fig:l and a}
\end{figure}

Using a batch size of 30, the encoder-decoder trained across 400 epochs. Fig.~\ref{fig:l and a} shows the progression of training and validation accuracy and loss through epochs. The encoder-decoder's loss plateaued for the training data, reaching between 0.5 and 0 for the training data. It overfit to a certain extent, as the validation data's loss began to increase. The encoder-decoder's final training accuracy was 92.8\%. On validation data, it had a peak accuracy of 71.4\%. 

\begin{figure}[h]
\centering
    \includegraphics[width=\linewidth]{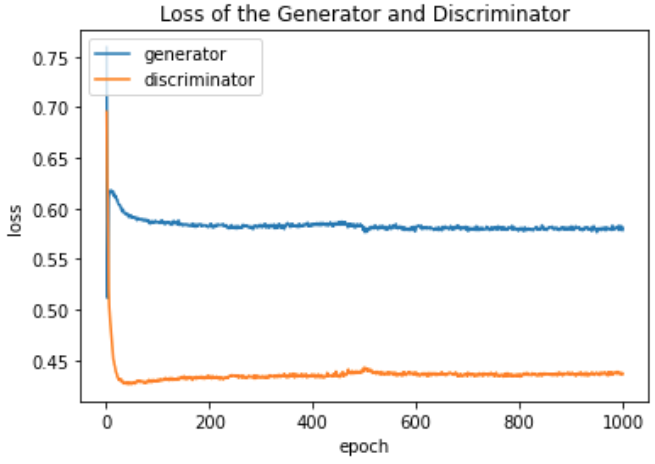}
    \caption{Loss of the GAN}
    \label{fig:ganloss}
\end{figure}

Using a batch size of 1900, the GAN trained across 8000 epochs. Fig.~\ref{fig:ganloss} shows the loss of the generator and the discriminator for the first 1000 epochs. The GAN’s loss values steeply dropped and reached a plateau for both the generator and the discriminator, indicating that the models reached convergence and were both performing optimally against each other. The next 3000 epochs were run to further refine the models' performance. The final loss values hovered around 0.581 for the generator and 0.438 for the discriminator. 

\section*{Acknowledgment}

Many thanks to Anu Datar and Ricky Grannis-Vu for their ongoing encouragement and support.

\bibliographystyle{IEEEtran}
\bibliography{bibtex/bib/custom, bibtex/bib/anthology}

\end{document}